\documentclass{article}

\PassOptionsToPackage{numbers, compress}{natbib}
  \usepackage[preprint]{neurips_2019}

\usepackage[utf8]{inputenc} %
\usepackage[T1]{fontenc}    %
\usepackage{hyperref}       %
\usepackage{url}            %
\usepackage{booktabs}       %
\usepackage{amsfonts}       %
\usepackage{nicefrac}       %
\usepackage{microtype}      %
\usepackage{tikz,pgfplots}
\usepackage{enumitem}
\usepackage{natbib}
\pgfplotsset{compat=newest}
\usepackage{color}
\definecolor{mblue}{HTML}{0992F2}
\definecolor{morange}{HTML}{ff7f0e}
\definecolor{mgreen}{HTML}{2ca02c}
\definecolor{mred}{HTML}{d62728}
\definecolor{mviolet}{HTML}{9467bd}
\definecolor{mbrown}{HTML}{8c564b}
\definecolor{mpink}{HTML}{e377c2}
\definecolor{mgray}{HTML}{7f7f7f}
\definecolor{myellow}{HTML}{bcbd22}
\definecolor{mcyan}{HTML}{17becf}

\definecolor{darkgreen}{rgb}{0,.5,0}
\definecolor{lightgreen}{rgb}{.5,.85,.5}

\usepackage{amsmath,amsthm,amssymb}
\usepackage{graphicx}

\graphicspath{{pics/}}

\newcommand{\permp}{\permpkml{k}{l}{m}}
\newcommand{\perm}{\permxkml{k}{l}{m}{\sigma}}
\newcommand{\permpkml}[3]{\permxkml{#1}{#2}{#3}{\theta}}
\newcommand{\permxkml}[4]{#4^{(#1)}_{#2\Leftrightarrow#3}}

\usepackage{xcolor}
\hypersetup{
    colorlinks,
    linkcolor={red!50!black},
    citecolor={blue!50!black},
    urlcolor={blue!80!black}
}

\title{Weight-space symmetry in deep networks gives rise to permutation saddles, connected by equal-loss valleys across the loss landscape}

\author{%
    Johanni Brea\thanks{Equal contribution} $^{1}$ \qquad Berfin Simsek$^{* 1,2}$ \qquad Bernd Illing$^{1}$ \qquad Wulfram Gerstner$^{1}$ \\
    \\
    $ {}^{ 1}$Laboratory of Computational Neuroscience (LCN) \\
    $ {}^{ 2}$Chair of Statistical Field Theory (CSFT) \\
    EPFL, Lausanne, Switzerland\\
    \\
  \texttt{\{johanni.brea, berfin.simsek, bernd.illing, wulfram.gerstner\}@epfl.ch} \\
}

\begin{document}

\maketitle

\begin{abstract}
  The permutation symmetry of neurons in each layer of a deep neural network gives rise not only to multiple equivalent global minima of the loss function, but also to first-order saddle points located on the path between the global minima.
  In a network of $d-1$ hidden layers with $n_k$ neurons in layers $k = 1, \ldots, d$, we construct smooth paths between equivalent global minima that lead through a `permutation point' where the input and output weight vectors of two neurons in the same hidden layer $k$ collide and interchange.
  We show that such permutation points are critical points with at least $n_{k+1}$ vanishing eigenvalues of the Hessian matrix of second derivatives indicating a {\em local} plateau of the loss function.
  We find that a permutation point for the exchange of neurons $i$ and $j$ transits into a flat valley (or generally, an {\em extended} plateau of $n_{k+1}$ flat dimensions) that enables all $n_k!$ permutations of neurons in a given layer $k$ at the same loss value.
  Moreover, we introduce high-order permutation points by exploiting the recursive structure in neural network functions, and find that the number of $K^{\text{th}}$-order permutation points is at least by a factor $\sum_{k=1}^{d-1}\frac{1}{2!^K}{n_k-K \choose K}$ larger than the (already huge) number of equivalent global minima.
  In two tasks, we illustrate numerically that some of the permutation points correspond to first-order saddles (`permutation saddles'): first, in a toy network with a single hidden layer on a function approximation task and, second, in a multilayer network on the MNIST task.
  Our geometric approach yields a lower bound on the number of critical points generated by weight-space symmetries and provides a simple intuitive link between previous mathematical results and numerical observations.
\end{abstract}

\section{Introduction}

In a multilayer network of $d-1$ hidden layers with $n$ neurons each, there are $(n!)^{d-1}$ equivalent configurations corresponding to the permutation of neuron indices in each layer of the network \citep{Goodfellow16, Bishop95}. The permutation symmetries give rise to a loss function (`loss landscape’) where any given global minimum in the weight space must have $(n!)^{d-1} - 1$ completely equivalent `partner’ minima.
Since the structure of the loss landscape plays an important role in the optimization of neural network parameters, a large number of numerical
\citep{Dauphin14, Goodfellow14, Li18, Sagun14, Sagun16} and mathematical \citep{Choromanska15, Rasmussen03, Freeman16} studies have explored the properties of the loss landscape.
However, only very few authors \cite{Fukumizu2000, Saad95} have
considered the influence of symmetries on the loss landscape.

\begin{figure}[h]
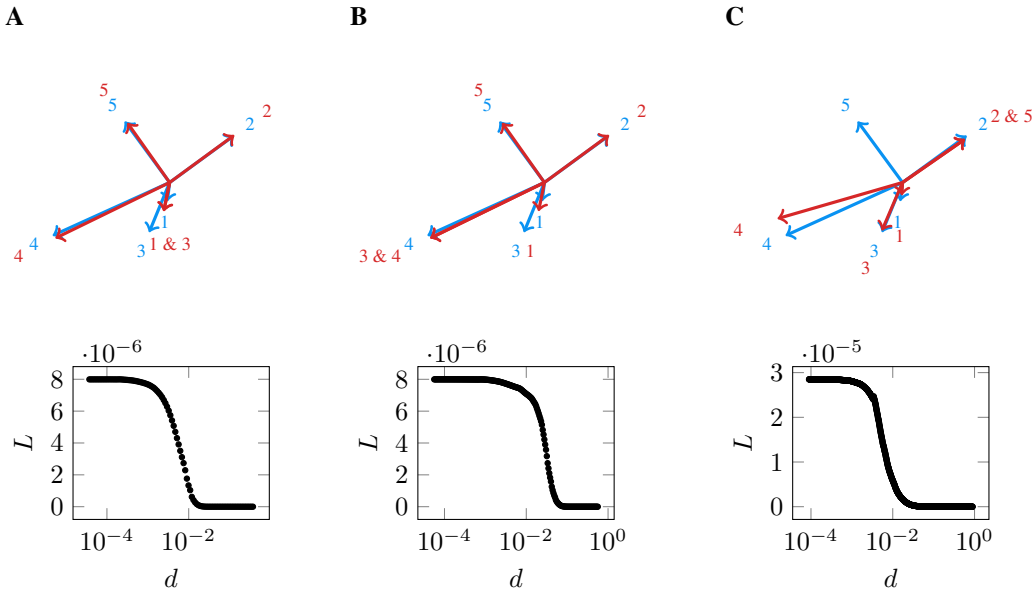

    \centering
\begin{tabular}{lll}
        \textbf{A} &\textbf{B} &\textbf{C} \\
\end{tabular}
    \caption{\textbf{Paths to a permutation point in networks with two-dimensional input space, five hidden neurons and one linear output neuron.}
    Merging hidden neurons $1 \& 3$ (A) leads to the same configuration and the same loss $L$ as merging neurons $3 \& 4$ (B) whereas merging neurons $2 \& 5$ leads to a different configuration and higher loss (C).
    \textbf{Top row}. Configuration  of the 5 weight vectors $W^{(1)}_{i,:}/b^{(1)}_i$ of the  hidden layer in a teacher network (blue) and at the permutation point (red).
Numbers indicate neuron indices.
\textbf{Bottom row}. Quadratic loss $L$ as a function of the distance $d$ between the units to be merged.
The distance was decreased in 200 logarithmically spaced steps from $d=d^{(1)}_{m,l}(\theta^*)$ to $1/10^4$ of the initial value. (The student network in the initial configurations has the parameters $\theta^*$ of the teacher network.)
For each $d$, full batch gradient descent on the loss $L$ was performed until convergence.
Training data was generated by sampling $10^3$ two-dimensional input points $x^\mu$ from a standard normal distribution and computing labels $y^\mu = f(x^\mu; \theta^*)$ using the teacher network. Teacher and student had a single layer of five hidden neurons with rectified-linear activation function $g$ and one linear output layer.
}\label{fig:toyexample}
\end{figure}

We wondered whether we can advance our understanding of the loss landscape, in particular the number of saddle points \cite{Dauphin14} and the extension of plateaus \cite{Goodfellow14}, by a careful analysis of weight-space symmetries of neural networks.
We start from the known permutation symmetries and consider smooth paths that connect two equivalent global minima via a saddle point.
The paths can be constructed by including into the loss function a scalar constraint that controls the distance between input weight vectors of two neurons in the same layer (see \autoref{fig:toyexample}).
At the `permutation point' where
the distance between the input weight vectors of the two neurons vanishes and their output weight vectors are identical, the indices of the two neurons can be interchanged at no extra cost and, after the change, the system returns on the same path back to the original configuration - except for the permutation of one pair of indices.

Surprisingly, we find that
the permutation point reached by moving along the path that merges a {\em single} pair of neurons allows us to permute {\em all} neuron indices in the same layer at the same cost.
These constant-loss permutations are possible because each permutation point lies in a subspace of
critical points with numerous flat directions - likely to correspond to the 'plateaus' that have been observed in numerical studies, e.g., \cite{Dauphin14, Goodfellow14}.
Our theory can be extended to higher-order saddles and provides explicit lower bounds for the number of first-order, second-order, and third-order permutation points with obvious generalizations to higher-order saddles.
Numerically, we confirm the existence of first-order permutation saddles with the above properties.

The specific contributions of our work are:
(i) A simple low-loss path-finding algorithm linking partner global minima via a permutation point,  implemented by minimization under a single scalar constraint.
(ii) The theoretical characterization of the permutation points.
(iii) A proof that all permutations of neurons in a given layer can be performed at equal loss.
(iv) A lower bound for the number of first- and higher-order permutation points.
(v) Numerical demonstrations of the path finding method and of permutation points in multilayer neural networks.
\section{Related Work}
Despite the success of deep learning in diverse application domains, the
structure of the loss landscape is only partly understood. Explorations of the loss landscape
\citep{Dauphin14,Goodfellow14, Li18, Sagun14, Sagun16, Choromanska15, Rasmussen03, Freeman16, ballard2017} are often driven by the question of how
gradient descent achieves excellent training accuracy despite a
highly non-convex loss landscape
\citep{Zhang16, Goodfellow16}.

\paragraph{Structure of the loss landscape.}

One line of research suggests that the landscape is relatively easy. In this context, it was shown that all the critical points -- except for the global minimum --  are saddles in the case of two-layered \citep{Baldi89} or multilayer \citep{Freeman16, Kawaguchi16, Lu17} linear networks. Interestingly, deep linear networks are reported to exhibit sharp transitions
at the edges of extended plateaus \citep{Saxe13}, similar to the plateaus observed in deep nonlinear networks \citep{Goodfellow14}.
For nonlinear multilayer networks, \citet{Choromanska15} argue that all local minima lie below a certain loss value by drawing connections to the spherical spin-glass model. Improving upon this result, recent theoretical work \cite{Soudry16, Nguyen17} shows that almost all local minima are global minima for multilayer networks under mild over-parametrization assumptions.

\paragraph{Bottom of the loss landscape.}

A second line of research studies the bottom of the landscape containing global minima and low-loss barriers between them.
For two-layered infinitely wide networks, global minima are connected with a path that is upper-bounded with a parameter that depends on the number of parameters in the network and data smoothness \citep{Freeman16}. Two numerical methods have been proposed to find sample paths between observed minima, either using a parameterized curve \citep{Garipov18} or by relaxation from linear paths with the Nudged Elastic Band method \citep{Draxler18}.

\paragraph{Moving down the loss landscape.}

A third line of research focuses on the question whether optimization paths may be slowed down close to, or even get stuck in, saddles and flat regions.
\citet{Lee16} show that gradient descent with sufficiently small step-size converges to local minima for general loss functions if all the saddles have at least one negative eigenvalue of the Hessian.
In addition, it is proven that gradient descent converges to global minima for infinitely wide multilayer networks \citep{Jacot18} and poor local minima are not encountered in over-parametrized networks \citep{spigler2018}.
Although these theoretical results guarantee convergence to non-saddle critical points, no guarantees are given for the speed of convergence. For soft-committee machines, it turns out that the initial learning dynamics is slowed down by correlation of hidden neurons \citep{Saad95, Engel01, Inoue2003}.
Using a projection method, \citet{Goodfellow14} find that the energy landscape contains plateaus with several flat directions, but report that optimization paths avoid these plateaus. \citet{Dauphin14} empirically argue that the large number of saddle points in the landscape makes training slow. \citet{sagun2017} claim that training is slowed down at the bottom of the landscape due to connected components and \citet{Baity18} attribute the empirically observed slow regime in the training to the large number of flat directions.

In this paper we show that there is an impressively large number of permutation points. Each permutation point is a critical point (either a local minimum or a saddle) with a large number of flat directions, potentially linked to the empirically observed plateaus.
In contrast to an earlier study by \citet{Fukumizu2000} with a scalar output for two-layered networks where a line of critical points around the permutation point was reported, we study a deep network with $d-1$ hidden layers and find multi-dimensional equal-loss plateaus. Moreover we give a novel lower bound on the number of permutation points and construct sample paths between global minima using an algorithm that is different from previously used methods \citep{Garipov18, Draxler18} since it exploits the symmetries at the permutation point.

\section{Preliminaries}

We study multilayer neural networks $f(x; \theta)$ with input $x\in \mathbb{R}^{n_0}$, $d$ layers of $n_1, \ldots, n_{d}$ neurons per layer, parameters  $\theta = \{ W^{(k)} \in \mathbb{R}^{n_k\times n_{k-1}} \mbox{ and } b^{(k)} \in \mathbb R^{n_k}:k=1,\ldots, d \}\in\Theta$ and $n_d$-dimensional output
\begin{equation}
    f(x; \theta) = {W}^{(d)}g\bigg(\cdots g\Big({W}^{(2)}g\big({W}^{(1)}x + b^{(1)}\big) + b^{(2)}\Big)\cdots\bigg) + b^{(d)}\, ,
\end{equation}
where $g$ is a nonlinear activation function that operates componentwise on any vector.
Since one can permute the neurons within each layer without changing the network function $f(x; \theta)$, any point $\theta$ induces a \emph{permutation set}
\begin{equation}
    P(\theta) = \{\theta'\in\Theta : W'^{(k)}_{\sigma^{(k)}(i),\sigma^{(k-1)}(j)} = W^{(k)}_{i,j}\mbox{ and } b'^{(k)}_{\sigma^{(k)}(i)} = b^{(k)}_i, k = 1,\ldots,d\}\, ,
\end{equation}
of points $\theta'$ with $f(x; \theta) = f(x; \theta')$, where $\sigma^{(k)}$ are permutations of the neuron indices $\{1, \ldots, n_k\}$ in (hidden) layer $k$ where $\sigma^{(0)}$ and $\sigma^{(d)}$ are fixed trivial permutations, since we do not want to permute neither the indices of the input nor that of the output.
We will use the notation $\theta' = \perm(\theta)$ to indicate a point $\theta'$ that differs from $\theta$ only by swapping neurons $l$ and $m$ in layer $k$ and $\vartheta^{(k)}_m = \big(W^{(k)}_{m,1},\ldots,W^{(k)}_{m,n_{k-1}}, b^{(k)}_m\big)$
to denote the parameter vector of neuron $m$ in layer $k$.
Note that the cardinality of a permutation set is maximal $|P(\theta) | = \prod_{j=1}^{d-1}n_j!$ only if all parameter vectors $\vartheta_l^{(k)}\neq \vartheta_m^{(k)}$ are distinct for every $l\neq m$ and layer $k=1,\ldots,d-1$. In the following,
we will assume that, at global minima, all parameter vectors are distinct at every layer $k$.

For training data $D = \{(x^\mu, y^\mu): \mu = 1,\ldots, T\}$ with targets $y^\mu\in\mathcal Y$, we define a loss function $L(\theta; D) = \frac1T\sum_{\mu = 1}^T\ell\big(y^\mu, f(x^\mu; \theta)\big)$, where $\ell: \mathcal{Y}\times\mathbb{R}^{n_d} \to \mathbb R$ is some single sample loss function.
To simplify notation we will usually omit the explicit mentioning of the data in the loss function, i.e. $L(\theta)\equiv L(\theta; D)$.

\section{Main results}
\subsection{A method to construct continuous low-loss permutation paths}\label{sec:lowlosspath}

\begin{figure}[h]
  \centering
  \hbox{A \hspace{60mm}B}
  \hbox{
    \includegraphics[width=7cm]{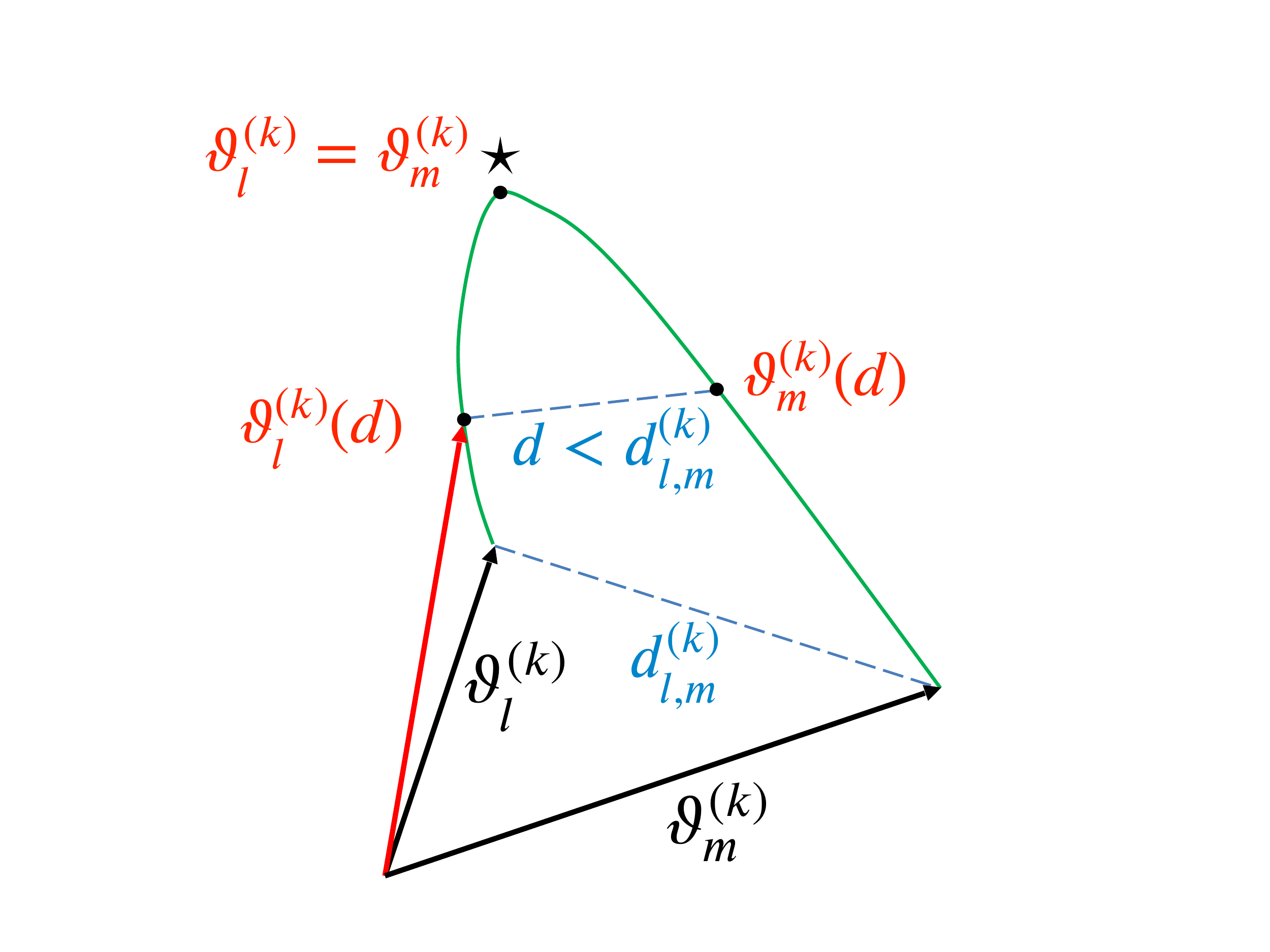}%
    \includegraphics[width=7cm]{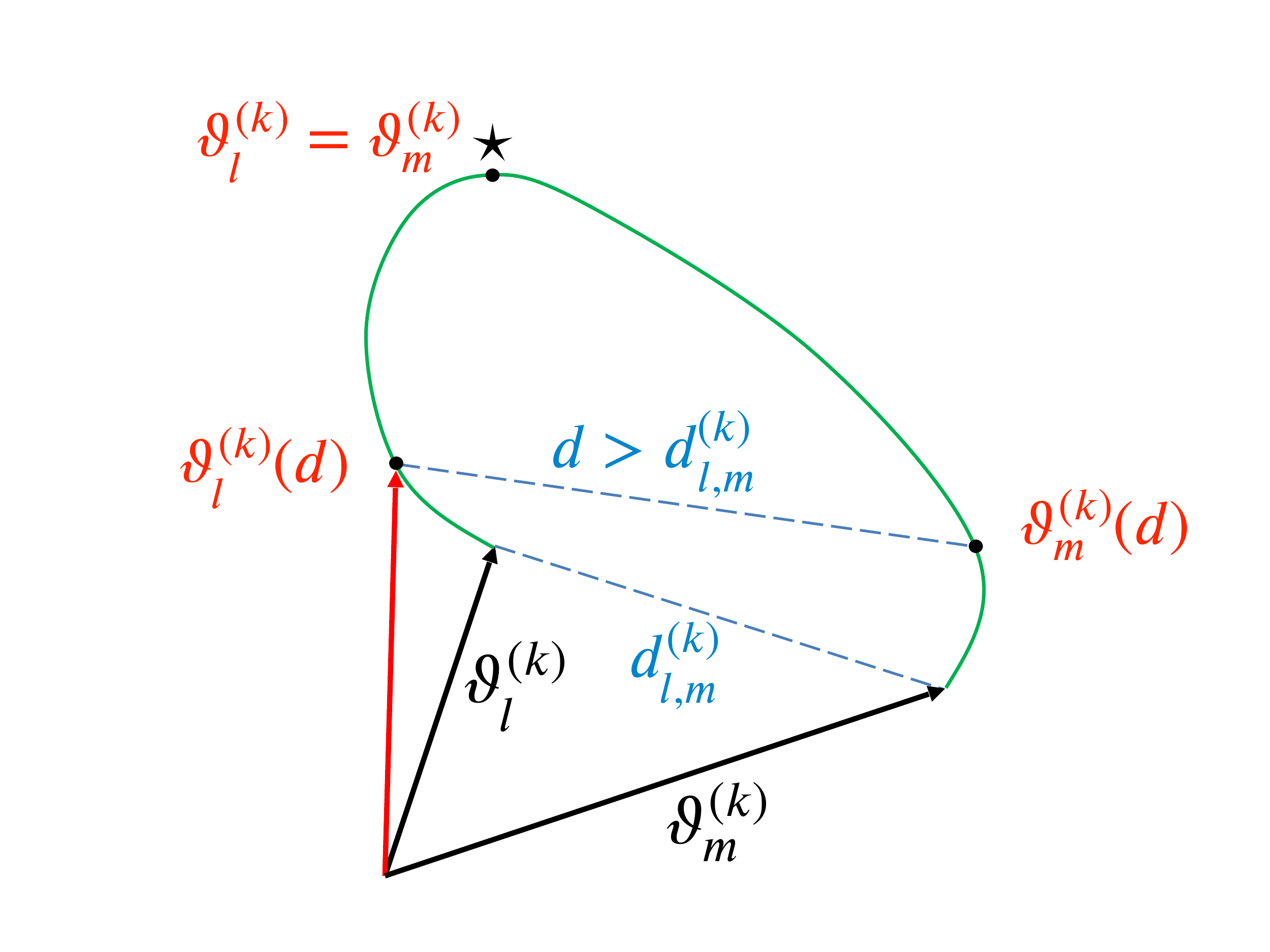}}
  \caption{{\bf A}.
      Configuration of two parameter vectors $\vartheta^{(k)}_l$ and $\vartheta^{(k)}_m$
      in the teacher network (black) and
      a potential path (green) towards a permutation point $\permp$ ($\star$) in the student network (black filled circles: positions
      of the student parameter vectors along the path; sample vector shown in red).
      The path is parametrized by the distance $d$.
      Along the path the distance $d$ (blue dashed lines) decreases continuously starting at $d_{l,m}^{(k)}$.
      Note that the path can lead to a permutation point far away from the initial configuration, outside the linear subspace spanned by the parameter vectors $\vartheta^{(k)}_l$ and $\vartheta^{(k)}_m$.
      {\bf B}. Minimizing the loss function $L(\theta)$ at every step excludes a hypothetical path (green) where the distance along the path increases.}
          \label{fig:schematic}
\end{figure}

We are interested in a continuous path $\gamma(t)$ where $t\in [0,1]$ and $\gamma$ lives in the space of parameters $\Theta$.
The path should have low loss $L(\gamma(t))$ and connect two distinct points in a permutation set $P(\theta)$, i.e. $\gamma(0) = \theta$ and $\gamma(1) = \theta'$ for $\theta\neq\theta'$ and $\theta, \theta' \in P(\theta)$.
To construct such a path, we first move to a point where the parameter vectors of two neurons $l$ and $m$ in layer $k$ coincide (see \autoref{fig:schematic}).
Once the parameter vectors $\vartheta_l^{(k)}=\vartheta_m^{(k)}$ are identical, both neurons compute the same function.
We can therefore modify output weights, i.e. increase $W^{(k+1)}_{n,m}$
by an amount $\epsilon$ and decrease $W^{(k+1)}_{n,l}$ by the same amount smoothly until the two output weights are equal for all neurons $1\le n \le n_{k+1}$.
This parameter configuration will be called a \emph{permutation point}, if it has locally minimal loss in all free parameters.
At this point we can permute the two indices at no cost and without any jump in parameter values.
After the exchange of indices we can walk back on the same path, leading back to the initial configuration of hyperplanes, but with one pair of indices permuted.

Formally, with $d^{(k)}_{l, m}(\theta) = \|\vartheta^{(k)}_l - \vartheta^{(k)}_m\|$ the Euclidean distance between the parameter vectors of neuron $l$ and $m$ in layer $k$ of configuration $\theta$, we can construct permutation paths with the following properties.
First, $d^{(k)}_{l, m}(\gamma(t)) = (\frac14 - t)d^{(k)}_{l, m}(\theta)$ for $t\in[0,\frac14]$,
i.e. the distance between the parameter vectors of neuron $l$ and $m$ in layer $k$ decreases continuously until it is zero at $t=\frac14$; for $t \in [\frac14,\frac12]$ the output weights of neurons $l$ and $m$ move at constant loss towards equality until a permutation point $\permp=\gamma(\frac12)$ is reached.
Second, $\gamma(t) = \perm\big(\gamma(1-t)\big)$ for $t\in[\frac12, 1]$, i.e. after reaching the permutation point the path continues in reversed order with neuron indices $l$ and $m$ of layer $k$ exchanged until one arrives at $\gamma(1) = \perm(\theta)$.
Third, the gradient $\nabla L(\gamma(t))$ is parallel to the tangent vector $\dot\gamma(t)$ and the Hessian $\nabla^2 L(\gamma(t))$ has at most one negative eigenvalue for all $t\in[0, 1]$.

If $\theta$ is a minimum of $L$, there must be a saddle point on such a path $\gamma$, potentially but not necessarily, at
the permutation point.
Since we cannot exclude that the Hessian has vanishing eigenvalues, the saddle point is of `weak first order' in the following sense: the gradient vanishes, the curvature in all but one direction is non-negative, and the curvature in one direction is negative.
We cannot exclude that there are several saddle points on this path.
Moreover, there is no guarantee that the highest saddle should be the one at the permutation point (see Supplementary Material Fig.~1).

To find such paths algorithmically, we reparametrize $\vartheta^{(k)}_m = \vartheta^{(k)}_l + \delta e$ where $\delta$ is a positive scalar and $e$ is a unit-length vector.
We start with $\delta = d^{(k)}_{l, m}(\theta)$.
Next, we decrease $\delta$ infinitesimally and perform gradient descent on the loss $L$ until convergence, where $\delta$ is fixed but all other parameters can change, including $\vartheta^{(k)}_l$ and $e$.
This is repeated until $\delta = 0$ and a permutation point is finally reached by shifting the respective output weights to the same value at equal loss (which can be done by keeping the sum of the output weights constant, see Supplementary Material Fig.~2).

\subsection{Characterization of permutation points \label{subsec:charpermpoints}}

Since the loss is minimized in all free parameters at permutation points, they are critical points of a network with one neuron less and have therefore the following known properties.
\begin{itemize}[leftmargin=1.4em]
    \item Permutation points $\permp$ of neurons $l$ and $m$ in layer $k$ are critical points of the original loss function, i.e.  the gradient $\nabla L\big(\permp\big) = 0$ (see \cite{Fukumizu2000}, Theorem 1).
    \item Permutation points $\permp$ are local minima or (weak) first-order saddle points, i.e. the Hessian $\nabla^2 L\big(\permp\big)$ has at most one negative eigenvalue (see \cite{Fukumizu2000}, Theorem 3).
\end{itemize}
Furthermore, we find the following properties.

(i) At permutation points $\permp$ the Hessian $\nabla^2 L\big(\permp\big)$ has at least $n_{k+1}$
  vanishing eigenvalues. Moreover, the permutation point lies in a $n_{k+1}$-dimensional space ('plateau') of critical points. If layer $k+1$ has a single output neuron, the plateau reduces to a line of critical points \cite{Fukumizu2000}.

(ii) All other permutations of neuron indices in layer $k$ can be performed by smooth equal-loss transformations starting from permutation points $\permp$ of neurons $l$ and $m$, i.e. there is a smooth path $\gamma:[0, 1]\rightarrow \Theta$ such that $\gamma(0) = \permp$ and $\gamma(1) = \permpkml{k}{i}{j}$ and $L(\gamma(t)) = L(\gamma(0))$ for all $t\in[0, 1]$ and all $i, j\in\{1, \ldots, n_k\}$.

      {\bf Proof sketch.}
      (i) Once the parameter vectors of neurons $m$ and $l$ in layer $k$
      are identical (which happens  at $t=0.25$ in our construction of the permutation paths), they implement the same function and, extending theorem 1 of \cite{Fukumizu2000}, achieve criticality.
      Any change of an output weight $W^{(k+1)}_{n,m}$ preserves the network function, and criticality, as long as the
      output weight
      $W^{(k+1)}_{n,l}$ is re-adapted so as to keep the sum
      $W^{(k+1)}_{n,m} + W^{(k+1)}_{n,l} $ constant.
       The sum-constraint $W^{(k+1)}_{n,m} + W^{(k+1)}_{n,l} = c $ for each $n$ in layer $k+1$ defines a $n_{k+1}$-dimensional hyperplane of critical points. In particular, at each point in the hyperplane,
       we have $n_{k+1}$ directions of the Hessian with zero Eigenvalues.

      (ii) We make the following sequence of continuous transformations that are all possible at fixed loss.
First, we  decrease the output weights of neuron $m$ to zero while increasing those of $l$ by the same amount,  keeping the sum of weights $W^{(k+1)}_{j,m} + W^{(k+1)}_{j,l}$ constant for each $j$ in layer $k+1$.
  Second, we change smoothly the input parameter vector of neuron $m$ to match those of an arbitrary other neuron $i$ in the same layer $k$ (see Supplementary Material Fig. 2).
  Third, we increase the output weights of neuron $m$ while decreasing those of neuron $i$ until all output weights of neuron $i$ are zero,  keeping the sum of weigths $W^{(k+1)}_{j,m} + W^{(k+1)}_{j,i}$ constant for each $j$ in layer $k+1$.
      Fourth, we reduce the input parameter vector of neuron $i$ to zero.
        Fifth, we increase the input parameter vector of neuron $i$ to match that of neuron $l$ at the permutation point.
          Finally, we equally share output weights between neurons $i$ and $l$ so that $i$ has the same weights as previously neuron $m$ at the permutation point.

Effectively, this procedure enables us to exchange an arbitrary neuron $i$ with neuron $m$, but the procedure can be repeated for further permutations. The permutations constructed in the proof of property (ii) start at permutation points where the parameter vectors of neurons $l$ and $m$ merge.
Therefore the loss associated with all the permutations constructed in the proof is $L(\permp)$, the one of this permutation point.
However, we could also begin with two other weight vectors $i$ and $j$ and construct the path leading to another $\permpkml{k}{i}{j}$, that has, in general, a different loss.
Repeating the above arguments starting at $\permpkml{k}{i}{j}$ also allows to exchange arbitrary indices of neurons in layer $k$. (q.e.d)

Therefore, each of the  $\frac{n_k (n_k-1)}{2}$ different permutation points of layer $k$ corresponds to a plateau of $n_{k+1}$ dimensions (and lies at the same time in a valley with respect to other dimensions) and this plateau  enables the exchange of all indices in layer $k$; some of these permutation points may have the same loss (see \autoref{fig:toyexample} A and B), others a different loss (see \autoref{fig:toyexample} B and C), depending
on the configuration of weight vectors in the student network with $n_k - 1$ neurons in layer $k$. %
Note that, amongst all these permutation points embedded in different plateaus, we could for example search for the one with the lowest cost -- and this lowest-cost permutation would then also connect all global minima caused by arbitrary permutations of neurons in layer $k$.

\subsection{Counting Higher-Order Permutation Points}

At a permutation point of a neural network with $(n_1, \ldots, n_d)$ neurons per layer, the  two (input) parameter vectors of neurons $l$ and $m$ in layer $k$ are the same ($\vartheta^{(k)}_l = \vartheta^{(k)}_m$) and so are their two vectors of output weights to layer $n_{k+1}$  (see \autoref{sec:lowlosspath}). We can now shift the summation of the output weights $W^{(k+1)}_{n,m} + W^{(k+1)}_{n,l}$ to one output weight, say $W^{(k+1)}_{n,m}$, and set the output weight $W^{(k+1)}_{n,l}$ to zero for each neuron $n$ at layer $k+1$. At this point, we can remove the neuron $l$ at layer $k$ without any change in the network function. Starting from this parameter configuration with one neuron removed, we can merge two further parameter vectors and remove a further neuron from layer $k$, and iterate until we have removed $K$ neurons. Alternatively, instead of removing $K$ neurons, we can use the same sequence of merging the (input and output) weight vectors of a subset of neurons in layer $k$, but keep their number at $n_k$.
We define a \textit{$K^\text{th}$-order permutation point at layer $k$}
as the the parameter configurations of the full neural network with $(n_1, \ldots, n_k, \ldots, n_d)$ neurons that 'simulates' the smaller network with $(n_1, \ldots, n_k - K, \ldots, n_d)$ via merging of input and output weight vectors
of a subset of neurons in layer $k$.

We note that this permutation point is a critical point for a neural network with $(n_1, \ldots, n_k - K, \ldots, n_d)$ neurons. Therefore, it is also a critical point (a local minima or a first-order saddle) for a neural network with $(n_1, \ldots, n_k - K + 1, \ldots, n_d)$ neurons (see \cite{Fukumizu2000}, Theorem 3). Exploiting this recursion $K$ times, we observe that a $K^\text{th}$-order permutation point at layer $k$ is a critical point of a neural network with $(n_1, \ldots, n_k, \ldots, n_d)$ neurons -- possibly a saddle of order $K$ or less. We are interested in counting the number of $K^\text{th}$-order permutation points at layer $k$ that give rise to the same network function as -- and are reducible to -- a smaller network with $K$ neurons removed from layer $k$.

{\bf Proposition 1}.
For $K \le n_k/2$ and $k=1, \ldots, d-1$, let $T(K,n_k)$ denote the ratio of the number of $K^\text{th}$-order permutations points at layer $k$ to the number of global minima.

(i) For $K=1,2,3$,  we find $T(K, n_k)$ to be:
  \begin{itemize}
    \item $T(K=1, n_k) = {n_k-1 \choose 1}\frac{1}{2!}$
    \item $T(K=2, n_k) = {n_k-2 \choose 1}\frac{1}{3!} + {n_k-2 \choose 2}\frac{1}{2!^2}$
    \item $T(K=3, n_k) = {n_k-3 \choose 1}\frac{1}{4!} + {n_k-3\choose 2} \frac{1}{3!} + {n_k-3\choose 3} \frac{1}{2!^3} $
  \end{itemize}

(ii)  For $K\le n_k/2$, we find the bound $T(K, n_k) \geq {n_k-K \choose K}\frac{1}{2!^K}$.

{\bf Proof Sketch}.

We simulate a smaller neural network with the big network of
$(n_1, \ldots, n_k, \ldots, n_d)$ neurons across the $d$ layers.
We  assume that the small neural network has $n_k - K$ distinct parameter vectors at layer $k$ and $n_j$ distinct parameter vectors at other layers $j \neq k$.
A $K^\text{th}$-order permutation point (at layer $k$) of the big network implements all the $(n_1, \ldots, n_k - K, \ldots, n_d)$  parameter vectors of the small network and only these. Since the big networks has $n_k$ neurons in layer $k$ and the small network only $n_k-K$, the big network must reuse some of these parameter vectors of the smaller network several times. Therefore we count the number of permutations of indices to calculate $T(K, n_k)$. We start with $K=1$.

(i) At a first-order permutation point in layer $k$, we have
$l=n_k-1$ distinct parameter vectors $\{\vartheta^{(k)}_1, \vartheta^{(k)}_2, \ldots, \vartheta^{(k)}_l\}$ for a total of $n_k$ neurons.
Therefore two of the $n_k$ neurons must have the same parameter vector. (More formally, there is only one way to partition $n_k$ into $n_k - 1$ positive integers without respecting order and this unordered partition can be represented as $n_k = 2 + 1 + \ldots + 1$).
The shared parameter vector could be the first one, $\{\vartheta^{(k)}_1\}$, or the second one, ... or the last one, $\{\vartheta^{(k)}_l\}$.
Therefore there are ${n_k - 1 \choose 1}$ choices.
For each of these choices (say, we double the third parameter vector), we have $\frac{n_k!}{2!}$ permutations of indices of neurons in layer $k$.
If we include the permutations that are possible at all other layers, we have ${n_k - 1 \choose 1}\frac{1}{2!}\prod_{j=1}^{d-1}n_j!$  first-order permutation points (see Supplementary Material Fig.~3).

Since the cardinality of the permutation set induced by a global minimum is $|P(\theta) | = \prod_{j=1}^{d-1}n_j!$, we arrive at a ratio %
 $T(K=1, n_k) = {n_k - 1 \choose 1}\frac{1}{2!}$. Similar counting arguments for $K=2, 3$ can be found in the Supplementary Material.

(ii) For general $K$, we have $l = n_k - K$ distinct parameter vectors in the small network.
There are many ways to partition $n_k$ into $l$ positive integers without respecting order. Since we are interested in a lower bound, we  only consider the following unordered partition: $n_k = 2 + \ldots + 2 + 1 + \ldots + 1$, i.e.  we have $K$ duplicated parameter vectors and $n_k - 2K$ parameter vectors that appear  once. For this unordered partition, we have ${n_k - K \choose K}$ ways to choose the duplicated parameter vectors.  For each one of these choices, we can permute the neuron indices  in $\frac{n_k!}{2!^K}$ different ways. Including the permutations in other layers $j \neq k$, we end up with ${n_k-K \choose K}\frac{1}{2!^K}\prod_{j=1}^{d-1}n_j!$ points in the permutation set.
The number is a lower bound of $T(K,n_k)$, because  other unordered partitions of $n_k$ give rise to other $K^\text{th}$-order permutation points at layer $k$. (q.e.d.)

{\bf Lemma 1}.
For finite $K \in \mathbb{Z}^{+}$, $\frac{1}{2!^K}{n_k-K \choose K} \rightarrow c_Kn_k^{K} \text{ as } n_k \rightarrow \infty$ (see Supplementary Material for the proof with Stirling's formula).

Considering all the layers, we note that the number of permutation points of order $K$ is at least $\sum_{k=1}^{d-1}\frac{1}{2!^K}{n_k-K \choose K}$ times more than the global minima for $2K \leq \min_{k}n_k$. When one layer has large number of neurons (i.e., $n_k \rightarrow \infty$) then the ratio $T(K, n_k)$ grows with $n_k^K$ (see Lemma 1).

We can generalize the property (i)  for the $1^{\text{st}}$-order permutation points (see \autoref{subsec:charpermpoints}) to the $K^{\text{th}}$-order permutation points.

{\bf Proposition 2}. A $K^\text{th}$-order permutation point at layer $k$ lies in a $Kn_{k+1}$-dimensional hyperplane of equal-loss parameter configurations (see Supplementary Material for the proof).

Let us consider a local minimum of the small network with $(n_1, \ldots, n_k - K, \ldots, n_d)$ neurons per layer. For $2<K \leq n_k/2$, the number of $K^\text{th}$-order permutation points at layer $k$ of the big network that allow to simulate the small network is huge (Proposition 1). Moreover each of these permutation points lies inside one of the hyperplanes (Proposition 2, see Supplementary Material Fig.~4).  Therefore, neural network landscapes exhibit numerous high-dimensional flat plateaus at various loss values and each one of these plateaus corresponds to a configuration of a smaller neural network.

\section{Empirical results}
\begin{figure}[h]
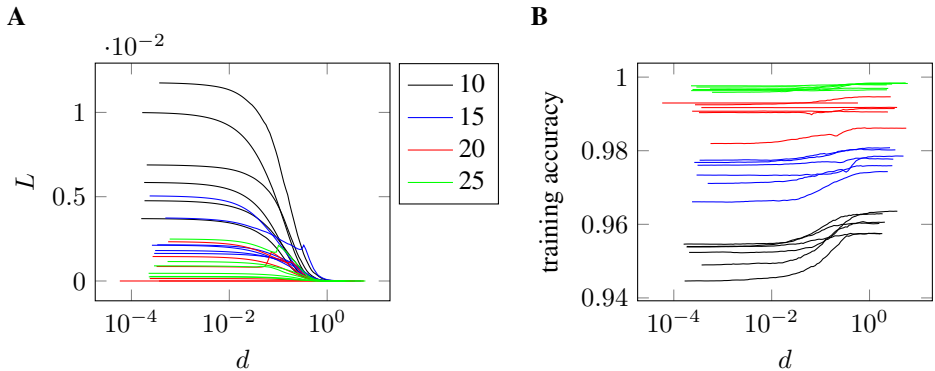

    \centering
\begin{tabular}{ll}
        \textbf{A} & \textbf{B} \\
\end{tabular}
    \caption{\textbf{A low-loss permutation path in the loss landscape of a network trained on MNIST}.
    \textbf{A}
    We merged two parameter vectors with high cosine-similarity in the second hidden layer of a three-layer network with $H=10, 15, 20$ or 25 trained on MNIST.
    For each hidden layer size we train 6 teacher networks with different random seeds and display one curve per hidden layer size and seed.
    The distance $d$ was decreased in 100 logarithmically spaced steps from $d^{(2)}_{m,l}(\theta^*)$ to $1/10^4$ of the original value.
    Full batch gradient descent was performed until convergence for every value of $d$.
    With $y^\mu = f(x^\mu; \theta^*) \in\mathbb{R}^{10}$ the output of the last layer before the softmax operation, we chose $L = \frac1T\sum_{\mu=1}^T\|y^\mu - f(x^\mu; \theta)\|^2/\langle {y_i^\mu}^2\rangle$, where $\langle .\rangle$ denotes the mean over patterns and dimensions.
    \textbf{B} The accuracy on the training set decreases only marginally when moving to a permutation point.
}\label{fig:mnist}
\end{figure}

Using a similar procedure as in the toy example (\autoref{fig:toyexample}), we constructed paths between global minima in fully connected three-layer network with $n_1=n_2=H$ and $n_3=10$ neurons (see \autoref{fig:mnist}).
We used a teacher-student setting: the teacher network was pre-trained on the MNIST data set using the negative log-likelihood loss function and its parameters were kept fixed thereafter. Then we replaced the softmax non-linearity of the last layer in the teacher network with the identity function and relabeled the original MNIST data with the 10-dimensional real-valued output of this adapted teacher network as the new target for a regression task with mean-squared error loss. We initialized the student network with the parameters $\theta^*$ of the teacher and decreased the distance between two selected parameter vectors in layer $k=2$.

Whereas in a few cases the trajectory toward the permutation point passed through a saddle on  the way, in most cases the loss increased monotonically toward the permutation point, indicating that the permutation point is a saddle, and not a minimum. As expected from theoretical results \citep{Freeman16}, the barrier height (loss at saddle) decreased with the number $H$ of hidden neurons per layer.

\section{Discussion}
\label{sec:discussion}

The suprising training performance of neural networks despite their highly non-convex nature has been drawing attention to the structure of the loss landscape, i.e. global and local minima, saddle points, flat plateaus and barriers between the minima.

In this paper, we explored how weight-space symmetry induces saddles and  plateaus in the neural network loss landscape. We found that special critical points, so-called \emph{permutation points}, are embedded in high-dimensional flat plateaus. We proved that all permutation points in a given layer are connected with equal-loss paths, suggesting new perspectives on loss landscape topology.
We provided a novel lower bound for the number of first- and higher-order permutation points and proposed a low-loss path finding method to connect equivalent minima. The empirical validation of our path finding algorithm in  a multilayer network trained on MNIST showed that permutation points could indeed be reached in practice. Additionally, we observed that the loss at the permutation point (barrier) decreased with network size and thus confirmed \citet{Freeman16}'s findings for loss barriers between global minima.
High-dimensional flat regions around permutation points could be one of the causes of the empirically observed slow phases in training.

\section{Acknowledgement}
We thank Mario Geiger and Levent Sagun for many interesting discussions. We thank Valentin Schmutz and Cl\'ement Hongler for their valuable feedback, especially for the theory part.

\newpage
\appendix
\section{Supplementary Figures}

\begin{figure}[h]
    \centering
    \includegraphics[width=\textwidth]{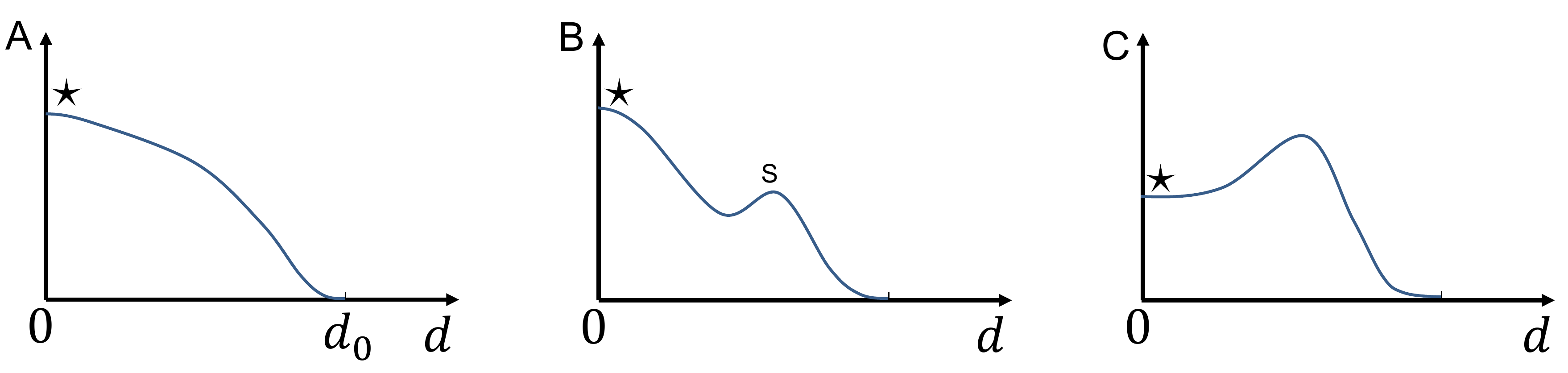}
    \caption{Loss $L$   (vertical axis) on the permutation path as a function of the distance $d$ between the two parameter vectors to be permuted (schematic).
      {\bf A}.
      In the teacher network the two parameter vectors have a distance $d_0$. Along the path, the distance is reduced to zero. At the permutation point ($\star$), the loss reaches a maximum which corresponds to a saddle point of the total loss function.
      {\bf B}. On the path towards the permutation point ($\star$) an intermediate saddle point (S) may occur.
      {\bf C}. The permutation point ($\star$) could be a minimum along the path.}
                      \label{fig-Loss-schematic1}
\end{figure}

\begin{figure}[h]
  \centering
   \hbox{A \hspace{10cm}B}
  \hbox{
  \includegraphics[width = 0.7\textwidth]{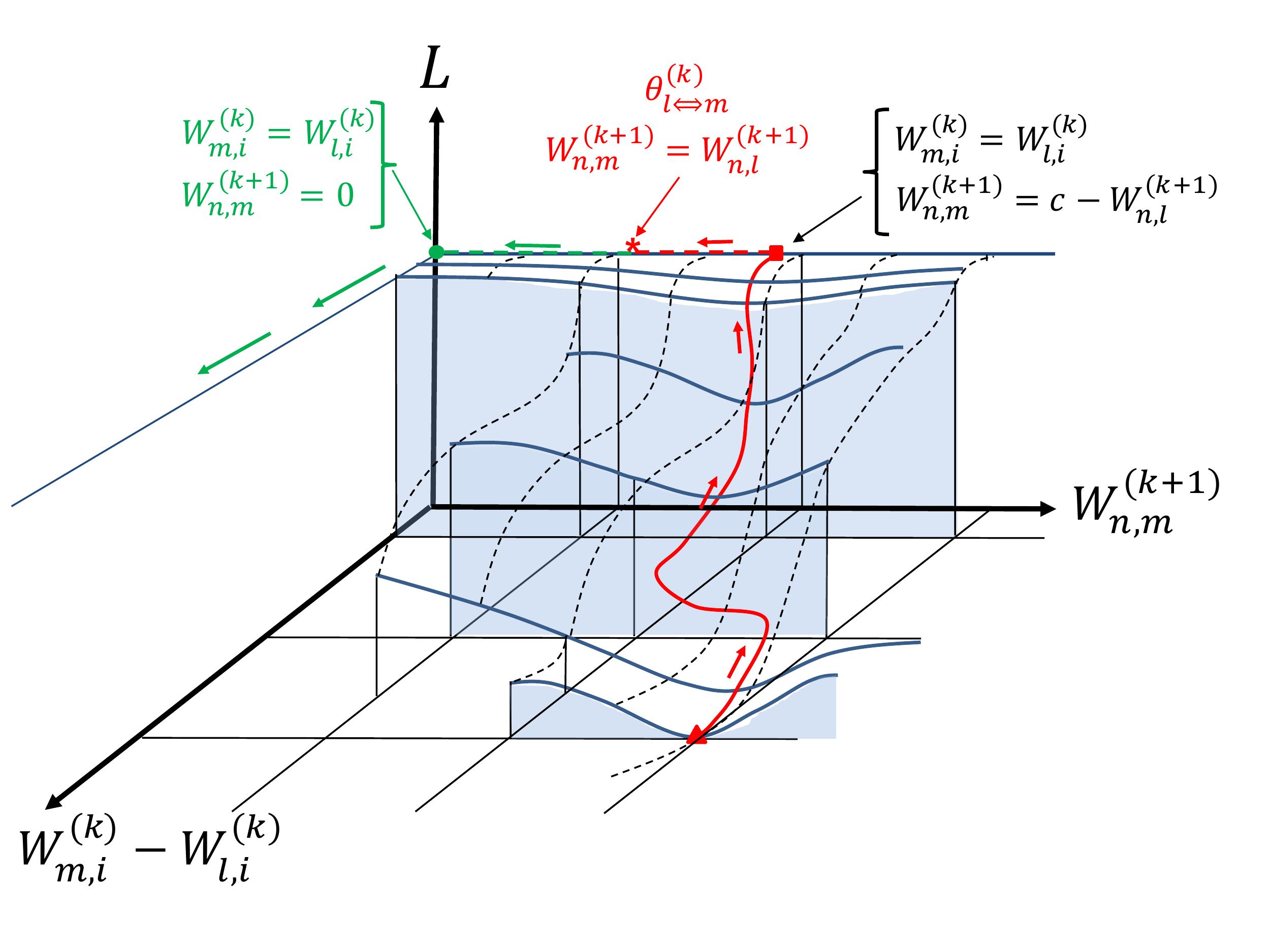}
  \includegraphics[width= 0.7\textwidth]{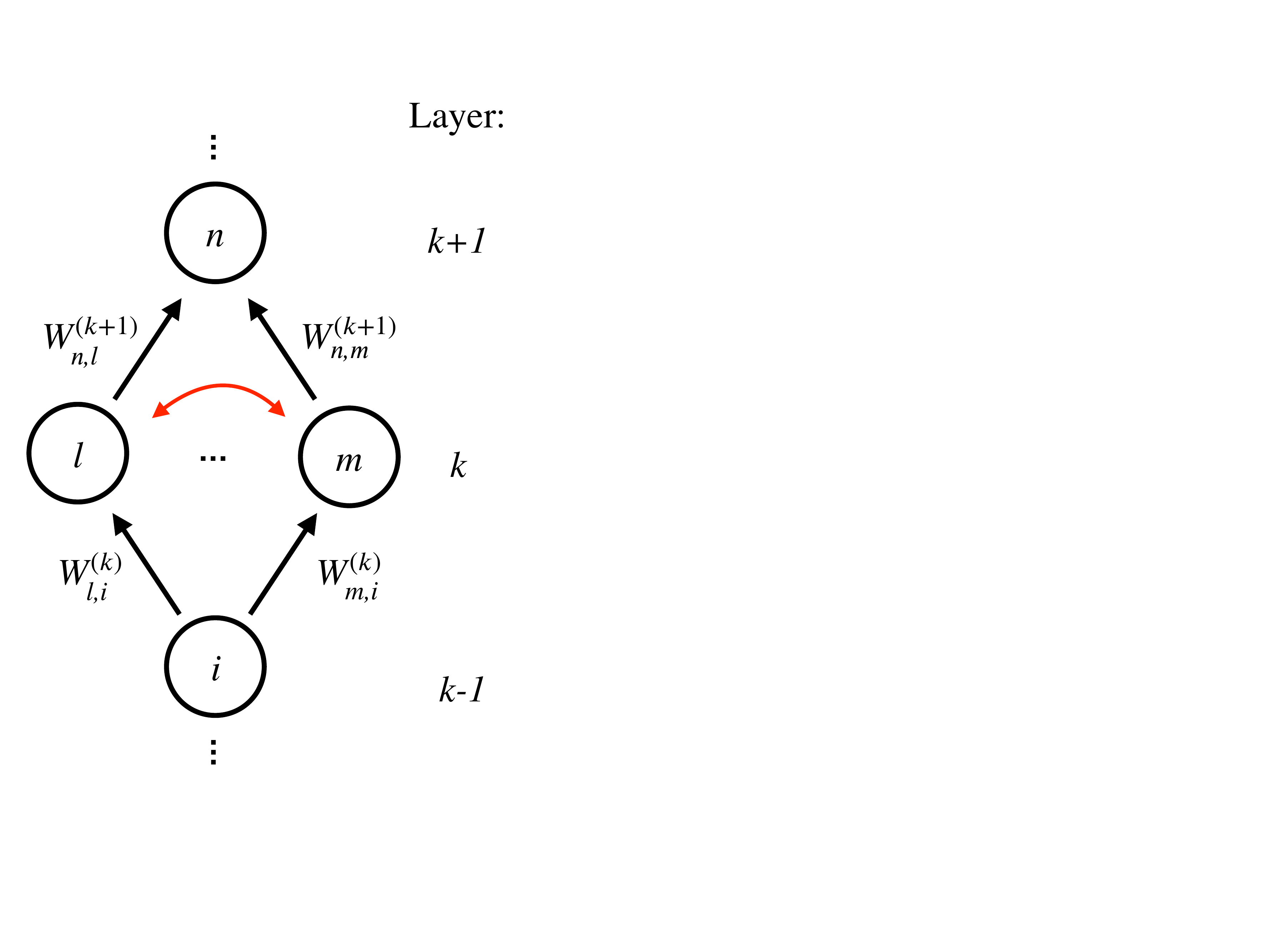}}
	\caption{{\bf A}. The loss landscape (schematic) as a function of two parameters: the
	difference $W_{m,i}^{(k)} - W_{l,i}^{(k)}$ between the weights from neuron
	$i$ to neurons $m$ and $l$ in layer $k$ and the
	weight $W_{n,m}^{(k+1)}$ from neuron $m$ to neuron $n$ in the next layer
	(see {\bf B} for network graph). The red curve indicates the path from one of the
	global minima (red triangle) to a saddle point where the difference
	between the input weight vectors of neurons $m$ and $l$ in layer $k$
	vanishes ($W_{m,i}^{(k)} = W_{l,i}^{(k)}$, red square). Along the axis $W_{m,i}^{(k)} - W_{l,i}^{(k)}=0$,
	we can change the output weight $W_{n,m}^{(k+1)}$ at constant loss
	(dashed horizontal line), as long as the sum
	$W_{n,m}^{(k+1)}+W_{n,l}^{(k+1)} = c$ remains constant. The point
	$W_{n,m}^{(k+1)} = W_{n,l}^{(k+1)}$ where the two output weights are
	identical (and assuming the same matching condition for the other
	output weights) defines the permutation point $\permp$ (red $\star$)
	where we can swap the indices of neurons $m$ and $l$ in layer $k$ at equal loss and continuously in parameter space.
	If we then shift (dashed green line) all output weights of neuron $m$ in layer $k$
	to zero ($W_{n,m}^{(k+1)}=0$ for all $n$, green filled circle), we are
	free to change the weight $W_{m,i}^{(k)}$ at constant loss (green arrows)
	so as to perform further permutations of neurons in layer $k$.}
\label{fig:landscape}
\end{figure}

\begin{figure}[h]
  \centering
  \hbox{ \hspace{1.5cm}
  \includegraphics[width = 0.7\textwidth]{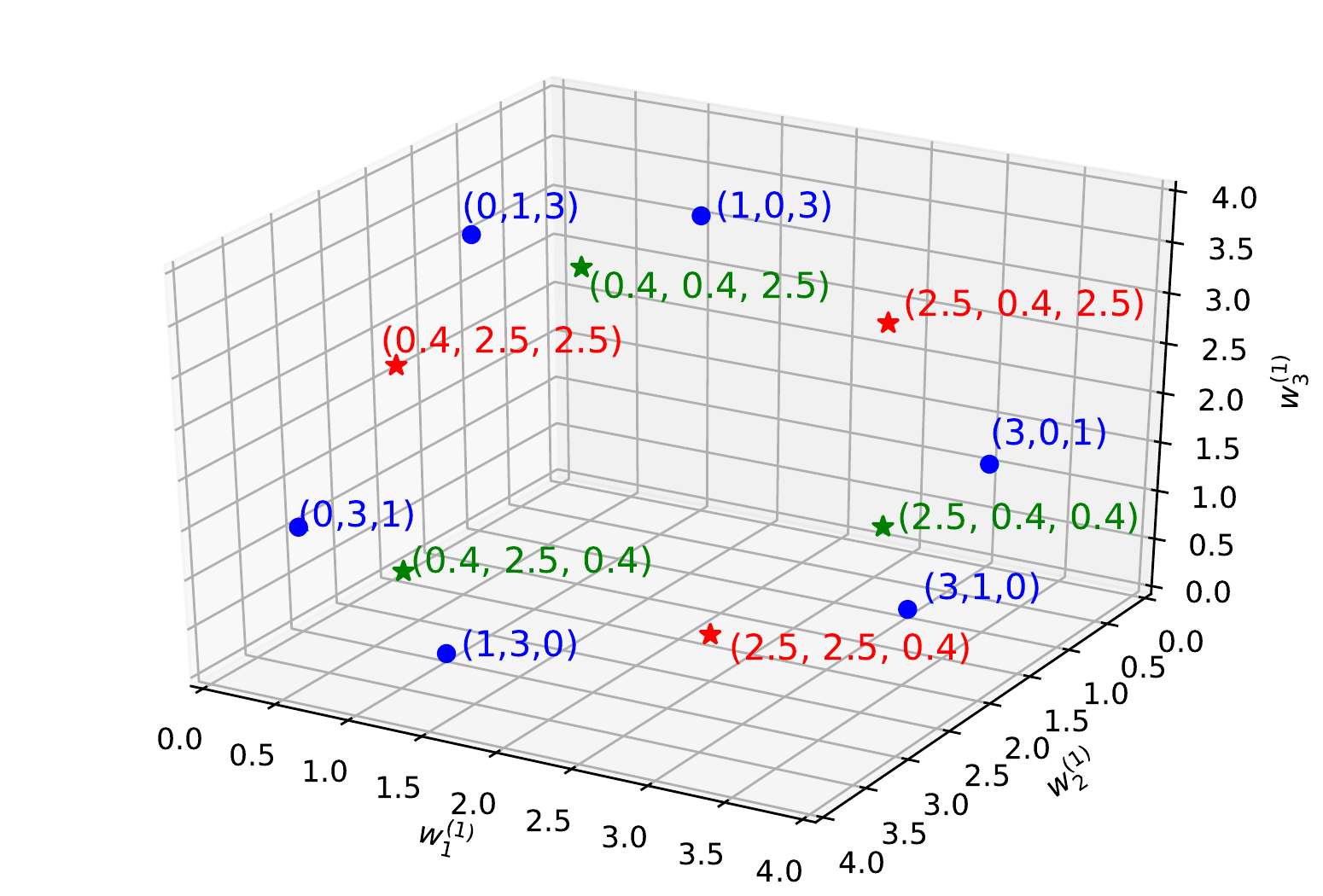}}
	\caption{Visualizing how permutation points arise in between global minima for a (hypothetical) network function with scalar input and one hidden layer with three neurons. Here we do not consider biases for simplicity. Blue dots: $3!$ equivalent global minima. A red $\star$: One permutation point (of first order at layer $k=1$) represented by its weights in the first layer, i.e. $(2.5, 2.5, 0.4)$. All red $\star$: One permutation set where the weight value $2.5$ is duplicated. All green $\star$: The other permutation set where the weight value $0.4$ is duplicated. Note that overall, we have $6$ permutation points that give rise to the same network function as the weight configuration with two hidden neurons with $(2.5, 0.4)$. Indeed, $T(K=1, n_1=3) = {3-1 \choose 1} \frac{1}{2!} = 1$ confirms why the number permutation points corresponding to this particular two hidden neurons weight configuration is equal to the number of global minima. Note that the weight values are assigned randomly for visualization purposes.}
\label{fig:counting1}
\end{figure}

\begin{figure}[h]
  \centering
  \hbox{A \hspace{7cm} B}
  \hbox{
  \includegraphics[width = 0.5\textwidth]{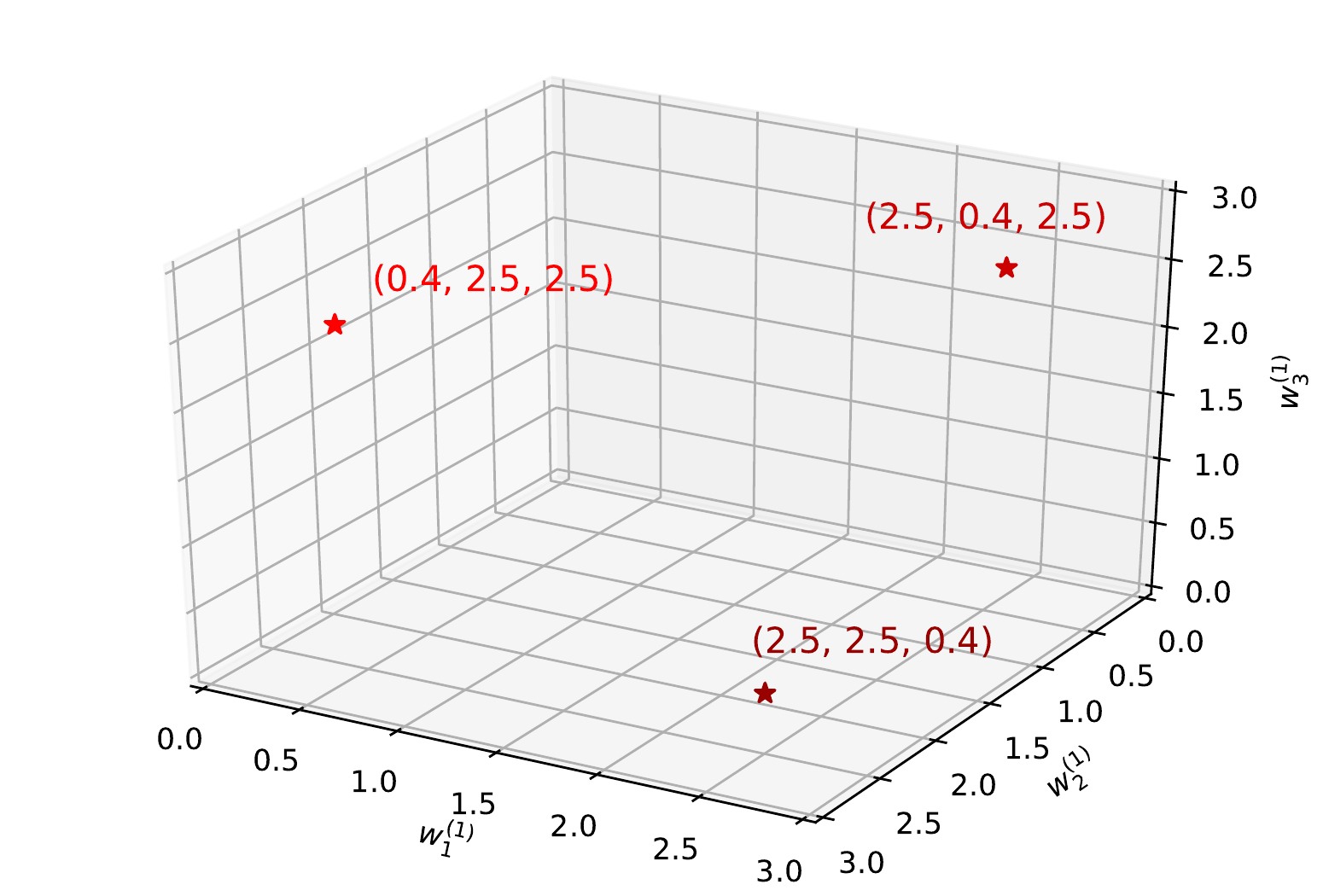}
  \includegraphics[width = 0.5\textwidth]{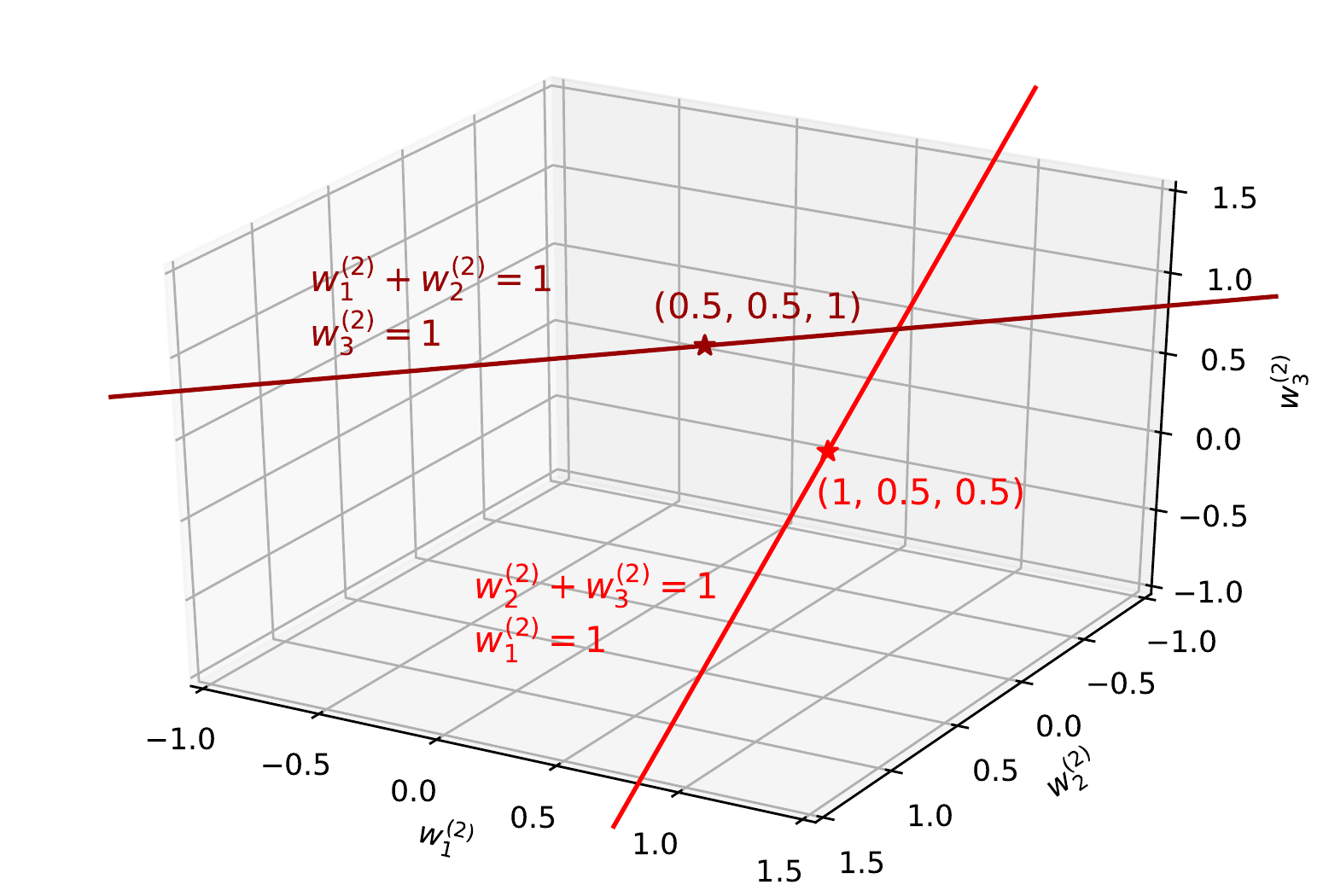}
  }
	\caption{{\bf A}. Zooming in permutations points of one of the permutation sets (red $\star$ in \autoref{fig:counting1}). {\bf B}. Visualizing how permutation points (at layer $k = 1$, see {\bf A}) lie in equal-loss lines in the weight space of the layer $k+1 = 2$. Only two out of three lines are shown for simplicity. Note that the number of such equal-loss lines (hyperplanes) is equal to the number of permutation points, i.e. each permutation point lies in one distinct line.}
\label{fig:counting2}
\end{figure}

\newpage ~ \newpage ~ \newpage

\section{Proof of Lemma 1}

{\bf Lemma 1.} For finite $K \in \mathbb{Z}^{+}$ and $n \in \mathbb{Z}^{+}$
  \begin{align}
    \ \frac{1}{2!^K}{n-K \choose K} \rightarrow c_Kn^{K} \text{ as } n \rightarrow \infty
  \end{align}

{\bf Proof.}

We can approximate the factorial an integer $n$ using Stirling's formula
\begin{align}
  \ n! \rightarrow \sqrt{2\pi n} \Big(\frac{n}{e}\Big)^n \text{ as } n \rightarrow \infty
\end{align}

We note that this approximation leads to accurate results even for small $n$.

As $n \rightarrow \infty$, both $n - K \rightarrow \infty$ and $n - 2K \rightarrow \infty$ for finite $K$. Therefore, we can apply Stirling's formula both for $n - K$ and $n - 2K$:
\begin{equation}
  \ \frac{1}{2!^K}{n - K \choose K} =  \frac{1}{2!^K} \frac{(n - K)!}{(n - 2K)!K!}
\end{equation}
\begin{equation}
  \ \frac{1}{2!^K} \frac{(n - K)!}{(n - 2K)!K!} \rightarrow \frac{1}{2!^K} \frac{\sqrt{2\pi (n-K)} \Big(\frac{n-K}{e}\Big)^{n-K}}{\sqrt{2\pi (n-2K)} \Big(\frac{n-2K}{e}\Big)^{n-2K} K!} \text{  as } n \rightarrow \infty
\end{equation}
\begin{equation}
  \ = \frac{1}{2!^K} \sqrt{\frac{n-K}{n-2K}} \frac{(n - K)^{n-K}}{(n - 2K)^{n-2K}} \frac{1}{e^K K!} \text{  as } n \rightarrow \infty
\end{equation}
\begin{equation}
  \ = \frac{1}{2!^K} \sqrt{\frac{\frac{n}{K}-1}{\frac{n}{K}-2}} \frac{(\frac{n}{K}-1)^{n-K}}{(\frac{n}{K}-2)^{n-2K}}K^K \frac{1}{e^K K!} \rightarrow \frac{1}{2!^K}(\frac{n}{K})^K \frac{K^K}{e^K K!}  \text{  as } n \rightarrow \infty
\end{equation}
\begin{equation}
  \ = \frac{1}{2!^K} \frac{1}{e^K K!} n^K  = c_K n^K \text{  as } n \rightarrow \infty
\end{equation}

\section{Counting $K^{\text{th}}$ order permutation points at layer $k$}

Following the counting arguments in the main text, we will focus on counting the number of permutations of parameter vectors at layer $k$.

\subsection{The case $K=2$}

There are two ways to have $n_k - 2$ distinct vectors out of $n_k$, corresponding to two unordered partitions of $n_k$: (i) $n_k = 3 + 1 + \ldots + 1$, and (ii) $n_k = 2 + 2 + 1 + \ldots + 1$. \\

(i) {\bf $n_k = 3 + 1 + \ldots + 1$} \\
For this case, we have $\frac{n_k!}{3!}$ permutations given by permuting the neuron indices of layer $k$ instead of the usual $n_k!$ permutations since we should eliminate the equivalent permutations corresponding to the permutations among the replicated parameter vectors with a division by $3!$. Therefore, this $2^\text{nd}$-order permutation point induces a permutation set with cardinality $|P(\theta)| = \frac{1}{3!}\prod_{j=1}^{d-1}n_j!$.

Now we should consider other $2^\text{nd}$-order permutation points (at layer $k$) giving rise to the same network function and corresponding to the same unordered partition, which we did not count in this permutation set. If we had chosen another parameter vector to replicate three times, this $2^\text{nd}$-order permutation point would induce another permutation set. Note that we can choose the parameter vector to replicate out of $n_k -2$ in ${n_k - 2 \choose 1}$ ways and there are $\frac{1}{3!}\prod_{j=1}^{d-1}n_j!$ many points in each one of the permutation sets. Therefore, we end up having ${n_k - 2 \choose 1}\frac{1}{3!}\prod_{j=1}^{d-1}n_j!$ many $2^\text{st}$-order permutation points (at layer $k$) corresponding to this unordered partition. \\

(ii) {\bf $n_k = 2 + 2 + 1 + \ldots + 1$} \\
For this case, we have $\frac{n_k!}{2!^2}$ permutations given by permuting the neuron indices of layer $k$ instead of the usual $n_k!$ permutations since we should eliminate the equivalent permutations corresponding to the permutations among the two pairs of duplicated parameter vectors with a division by $2!^2$. Therefore, this $2^\text{nd}$-order permutation point induces a permutation set with cardinality $|P(\theta)| = \frac{1}{2!^2}\prod_{j=1}^{d-1}n_j!$.

Again, we should consider other $2^\text{nd}$-order permutation points (at layer $k$) giving rise to the same network function and corresponding to the same unordered partition. Note that we can choose two parameter vectors to duplicate out of $n_k -2$ in ${n_k - 2 \choose 2}$ ways and there are $\frac{1}{2!^2}\prod_{j=1}^{d-1}n_j!$ many points in each one of the permutation sets. Therefore, we end up having ${n_k - 2 \choose 2}\frac{1}{2!^2}\prod_{j=1}^{d-1}n_j!$ many $2^\text{nd}$-order permutation points (at layer $k$) corresponding to this unordered partition. \\

Overall, we have ${n_k - 2 \choose 1}\frac{1}{3!}\prod_{j=1}^{d-1}n_j! + {n_k - 2 \choose 2}\frac{1}{2!^2}\prod_{j=1}^{d-1}n_j!$ many
$2^\text{nd}$-order permutation points at layer $k$.

\subsection{The case $K=3$}

There are three ways to have $n_k - 3$ distinct vectors out of $n_k$, corresponding to three unordered partitions of $n_k$: (i) $n_k = 4 + 1 + \ldots + 1$, (ii) $n_k = 3 + 2 + 1 + \ldots + 1$, and (iii) $n_k = 2 + 2 + 2 + 1 + \ldots + 1$. \\

(i) {\bf $n_k = 4 + 1 + \ldots + 1$} \\
 For this case, we have $\frac{n_k!}{4!}$ permutations given by permuting the neuron indices of layer $k$. Therefore, this $3^\text{rd}$-order permutation point induces a permutation set with cardinality $|P(\theta)| = \frac{1}{4!}\prod_{j=1}^{d-1}n_j!$.

As usual, we should consider other $3^\text{rd}$-order permutation points (at layer $k$) giving rise to the same network function and corresponding to the same unordered partition. If we had chosen another parameter vector to replicate four times, this $3^\text{rd}$-order permutation point would induce another permutation set. Note that we can choose the parameter vector to replicate out of $n_k -3$ in ${n_k - 3 \choose 1}$ ways and there are $\frac{1}{4!}\prod_{j=1}^{d-1}n_j!$ many points in each one of the permutation sets. Therefore, we end up having ${n_k - 3 \choose 1}\frac{1}{4!}\prod_{j=1}^{d-1}n_j!$ many $3^\text{rd}$-order permutation points (at layer $k$) corresponding to this unordered partition. \\

(ii) {\bf $n_k = 3 + 2 + 1 + \ldots + 1$} \\
 For this case, we have $\frac{n_k!}{3!2!}$ permutations given by permuting the neuron indices of layer $k$. Therefore, this $3^\text{rd}$-order permutation point induces a permutation set with cardinality $|P(\theta)| = \frac{1}{3!2!}\prod_{j=1}^{d-1}n_j!$.

As usual, we should consider other $3^\text{nd}$-order permutation points (at layer $k$) giving rise to the same network function and corresponding to the same unordered partition. If we had chosen two other parameter vector to replicate, say $\vartheta'^{(k)}_l$ and $\vartheta'^{(k)}_m$, this $3^\text{rd}$-order permutation point would induce another permutation set. Note that we can choose two parameter vectors to replicate out of $n_k -3$ in $2!{n_k - 3 \choose 2}$ ways. We have an extra $2!$ factor here since we have different permutation sets if we replicate $\vartheta'^{(k)}_l$ twice and $\vartheta'^{(k)}_m$ three times, or vice versa. Yet, there are $\frac{1}{3!2!}\prod_{j=1}^{d-1}n_j!$ many points in each one of the permutation sets. Therefore, we end up having $2!{n_k - 3 \choose 2}\frac{1}{3!2!}\prod_{j=1}^{d-1}n_j! = {n_k - 3 \choose 2}\frac{1}{3!}\prod_{j=1}^{d-1}n_j!$ many $3^\text{rd}$-order permutation points (at layer $k$) corresponding to this unordered partition. \\

(iii) {\bf $n_k = 2 + 2 + 2 + 1 + \ldots + 1$} \\ For this case, we have $\frac{n_k!}{2!^3}$ permutations given by permuting the neuron indices of layer $k$. Therefore, this $3^\text{rd}$-order permutation point induces a permutation set with cardinality $|P(\theta)| = \frac{1}{2!^3}\prod_{j=1}^{d-1}n_j!$.

As always, we should consider the other $3^\text{rd}$-order permutation points (at layer $k$) giving rise to the same network function and corresponding to the same unordered partition. Note that we can choose three parameter vectors to duplicate out of $n_k-3$ in ${n_k - 3 \choose 3}$ ways and there are $\frac{1}{2!^3}\prod_{j=1}^{d-1}n_j!$ many points in each one of the permutation sets. Therefore, we end up having ${n_k - 3 \choose 3}\frac{1}{2!^3}\prod_{j=1}^{d-1}n_j!$ many $3^\text{rd}$-order permutation points (at layer $k$) corresponding to this unordered partition. \\

Overall, we have ${n_k - 3 \choose 1}\frac{1}{4!}\prod_{j=1}^{d-1}n_j! + {n_k - 3 \choose 2}\frac{1}{3!}\prod_{j=1}^{d-1}n_j! + {n_k - 3 \choose 3}\frac{1}{2!^3}\prod_{j=1}^{d-1}n_j!$ many
$3^\text{rd}$-order permutation points at layer $k$.

\section{Proof of Proposition 2}

{\bf Proposition 2}. A $K^\text{th}$-order permutation point at layer $k$ lies in a $Kn_{k+1}$-dimensional hyperplane at equal-loss.

{\bf Proof}.

We will denote the parameter vectors of a neural network with $(n_1, \ldots, l=n_k-K, \ldots, n_d)$ neurons per layer with $\{\vartheta^{(k)}_1, \vartheta^{(k)}_2, \ldots, \vartheta^{(k)}_l\}$ and that of $f(n_1, \ldots, n_k, \ldots, n_d)$ neurons with $\{\vartheta'^{(k)}_1, \vartheta'^{(k)}_2, \ldots, \vartheta'^{(k)}_{n_k}\}$. Let's consider an unordered partition of $n_k = s_1 + s_2 + \ldots + s_l$ with $s_m \geq 1$ for $m=1, \ldots, l$. Without loss of generality, let's assume that the first $s_1$ parameter vectors of layer $k$ are the same, then the next $s_2$ and so on. Equivalently, for all $m=1, \ldots, l$, we have $\vartheta'^{(k)}_j = \vartheta^{(k)}_m \text{ for } j=s_{m-1}+1,s_{m-1}+2, \ldots,s_{m-1}+s_m$ where $s_0 = 0$.

The only free variables are the outgoing weights of the replicated neurons- except for the constraint that the summation of these weights is fixed. These constraints correspond to
$ \sum_{j=s_{m-1} + 1}^{s_{m-1} + s_m}W'^{(k+1)}_{\text{i,j}} = W^{(k+1)}_{\text{i,m}} \text{ for all neurons } i$ at layer $k+1$.
Overall, there are $(\sum_{m=1}^{l}s_m) n_{k+1} = n_{k}n_{k+1}$ free variables constrained by $ln_{k+1}$ equations. One equation defines a $n_{k}n_{k+1} - 1$ dimensional hyperplane in the
$n_{k} n_{k+1}$ space. Intersecting $ln_{k+1}$ of these hyperplanes, we end up having a $(n_{k} - l) n_{k+1} = K n_{k+1}$ dimensional equal-loss hyperplane.
\end{document}